\documentclass{article}

 \usepackage[main, final]{neurips_2025}

\usepackage[utf8]{inputenc} 
\usepackage[T1]{fontenc}    
\usepackage{hyperref}       
\usepackage{url}            
\usepackage{booktabs}       
\usepackage{graphicx}
\usepackage{amsfonts}       
\usepackage{amsmath}
\usepackage{tabularx}
\usepackage{multirow}
\usepackage{tablefootnote}
\usepackage{footnote}
\usepackage{nicefrac}       
\usepackage{microtype}      
\usepackage{xcolor}         
\usepackage{algorithm}
\usepackage{algpseudocode}
\usepackage{caption}
\usepackage{subcaption}

\usepackage{array} 

\newcolumntype{Y}{>{\centering\arraybackslash}X}

\title{FrèqFlow: Long-term forecasting using lightweight flow matching}

\author{%
  Seyed Mohamad Moghadas$^{1,2}$, Bruno Cornelis$^{1}$, Adrian Munteanu$^{1,2}\thanks{Senior IEEE Member}$\\
    $^{1}$ ETRO Department, Vrije Universiteit Brussel, B-1050 Brussels, Belgium \\
  $^{2}$ imec Kapeldreef 75, B-3001 Leuven, Belgium\\
  \texttt{Seyed.Mohamad.Moghadas@vub.be}, \texttt{bcorneli@etrovub.be}, \texttt{Adrian.Munteanu@vub.be} \\
}

\begin{document}

\maketitle



\begin{abstract}
Multivariate time-series (MTS) forecasting is fundamental to applications ranging from urban mobility and resource management to climate modeling. While recent generative models based on denoising diffusion have advanced state-of-the-art performance in capturing complex data distributions, they suffer from significant computational overhead due to iterative stochastic sampling procedures that limit real-time deployment. Moreover, these models can be brittle when handling high-dimensional, non-stationary, and multi-scale periodic patterns characteristic of real-world sensor networks. We introduce FrèqFlow, a novel framework that leverages conditional flow matching in the frequency domain for deterministic MTS forecasting. Unlike conventional approaches that operate in the time domain, FrèqFlow transforms the forecasting problem into the spectral domain, where it learns to model amplitude and phase shifts through a single complex-valued linear layer. This frequency-domain formulation enables the model to efficiently capture temporal dynamics via complex multiplication, corresponding to scaling and temporal translations. The resulting architecture is exceptionally lightweight with only $89k$ parameters—an order of magnitude smaller than competing diffusion-based models—while enabling single-pass deterministic sampling through ordinary differential equation (ODE) integration. Our approach decomposes MTS signals into trend, seasonal, and residual components, with the flow matching mechanism specifically designed for residual learning to enhance long-term forecasting accuracy. Extensive experiments on real-world traffic speed, volume, and flow datasets demonstrate that FrèqFlow achieves state-of-the-art forecasting performance, on average 7\% RMSE improvements, while being significantly faster and more parameter-efficient than existing methods. \href{https://github.com/moghadas76/Freq_Spect}{Github Repo}
\end{abstract}

\section{Introduction}
\label{sec:intro}

Forecasting multivariate time-series (MTS) is a foundational challenge in machine learning, critical for applications in urban mobility~\citep{Wen2023DiffSTG}, resource management~\citep{9666444}, and climate modeling \citep{Price2024, gao2025oneforecastuniversalframeworkglobal}. While recent generative models, particularly those based on denoising diffusion, have advanced the state-of-the-art in capturing complex data distributions \citep{NEURIPS2021_cfe8504b, Wen2023DiffSTG}, they often come with significant computational burdens. These models typically rely on iterative, stochastic sampling procedures that are computationally expensive and can be slow to converge, limiting their utility in real-time applications. Moreover, their performance can be brittle when faced with high-dimensional, non-stationary, and multi-scale periodic patterns characteristic of real-world sensor networks, such as urban traffic systems. An attractive alternative to stochastic diffusion is flow matching, a method for learning deterministic continuous-time transport maps between distributions \citep{lipman2023flowmatchinggenerativemodeling, feng2025guidanceflowmatching}. Instead of simulating a noisy diffusion process, flow matching models learn an Ordinary Differential Equation (ODE) that directly transports noise to data, enabling efficient, single-pass sampling~\citep{feng2025guidanceflowmatching}. Although its potential has been demonstrated in other domains, its application to multivariate spatio-temporal forecasting remains largely unexplored.

\begin{figure}[h]
  \centering
  \includegraphics[width=0.55\linewidth]{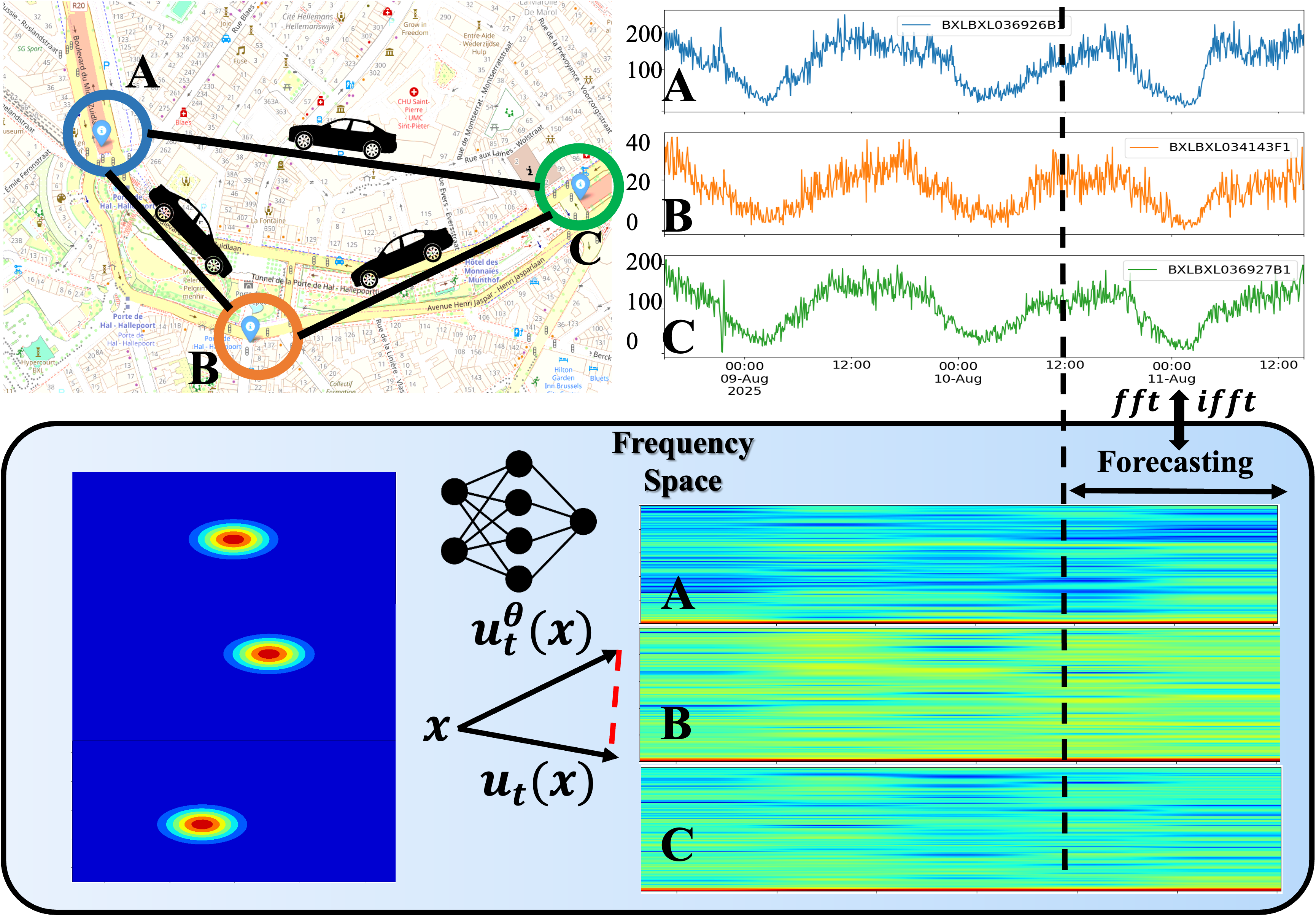}
  \caption{Problem setup and our proposed method, which is flow matching with a lightweight network in the frequency space. The underlying graph depicts Brussels’s road network in Belgium; the time-series signals are for the period \small{08-08-2025,14:55:00} for three days. Note that in the Node B, although the traffic pattern is highly correlated to the adjacent nodes, the traffic volume is significantly less.}
  \label{fig:my_figure_label}
\end{figure}

In this work, as illustrated by Figure~\ref{fig:my_figure_label}, we introduce FrèqFlow, a novel framework that integrates conditional flow matching within the frequency-domain space for MTS deterministic forecasting. Revealed by the research~\citep{li2025diffusionbased}, effective time-series forecasting by diffusion-based models is achievable through component-wise forecasting. Specifically, by decomposing a time-series into trend, seasonality, and residual components, they showed that diffusion blocks are effective for uncertainty modeling~\citep{li2025diffusionbased}. Inspired by this finding, we design our flow matching block for residual learning. Our key insight is that learning the velocity field in the frequency domain, rather than the time domain, allows for a more compact and efficient representation of complex temporal dynamics. By transforming the problem into the spectral domain, FrèqFlow learns to model amplitude and phase shifts, which correspond to scaling and temporal translations, respectively. This approach allows us to design a highly efficient architecture. The resulting model is exceptionally lightweight, comprising only $89k$ parameters, which is an order of magnitude smaller than many competing diffusion-based models. This compactness, combined with the single-pass nature of ODE-based sampling, leads to significant gains in inference speed without sacrificing forecasting accuracy. Our contributions are threefold:
\begin{itemize}
    \vspace{-0.17cm}
    \item To the best of our knowledge, we propose the first framework to leverage conditional flow matching in the frequency domain for MTS long-term deterministic forecasting, directly learning the velocity field of spectral components.
    \vspace{-0.17cm}
    \item We introduce a highly efficient, lightweight architecture that uses a single complex-valued linear layer to model temporal dynamics, drastically reducing computational cost.
    \vspace{-0.17cm}
    \item We demonstrate through extensive experiments on real-world traffic datasets that FrèqFlow achieves state-of-the-art or competitive performance while being significantly faster and more parameter-efficient than existing methods.
\end{itemize}

Our work bridges the gap between the generative power of continuous-time flows and the specific structural priors of spatio-temporal data, offering a practical and scalable solution for real-world traffic forecasting.

\section{Methodology}

\begin{figure}[h]
  \centering
  \includegraphics[width=0.8\linewidth, keepaspectratio=true]{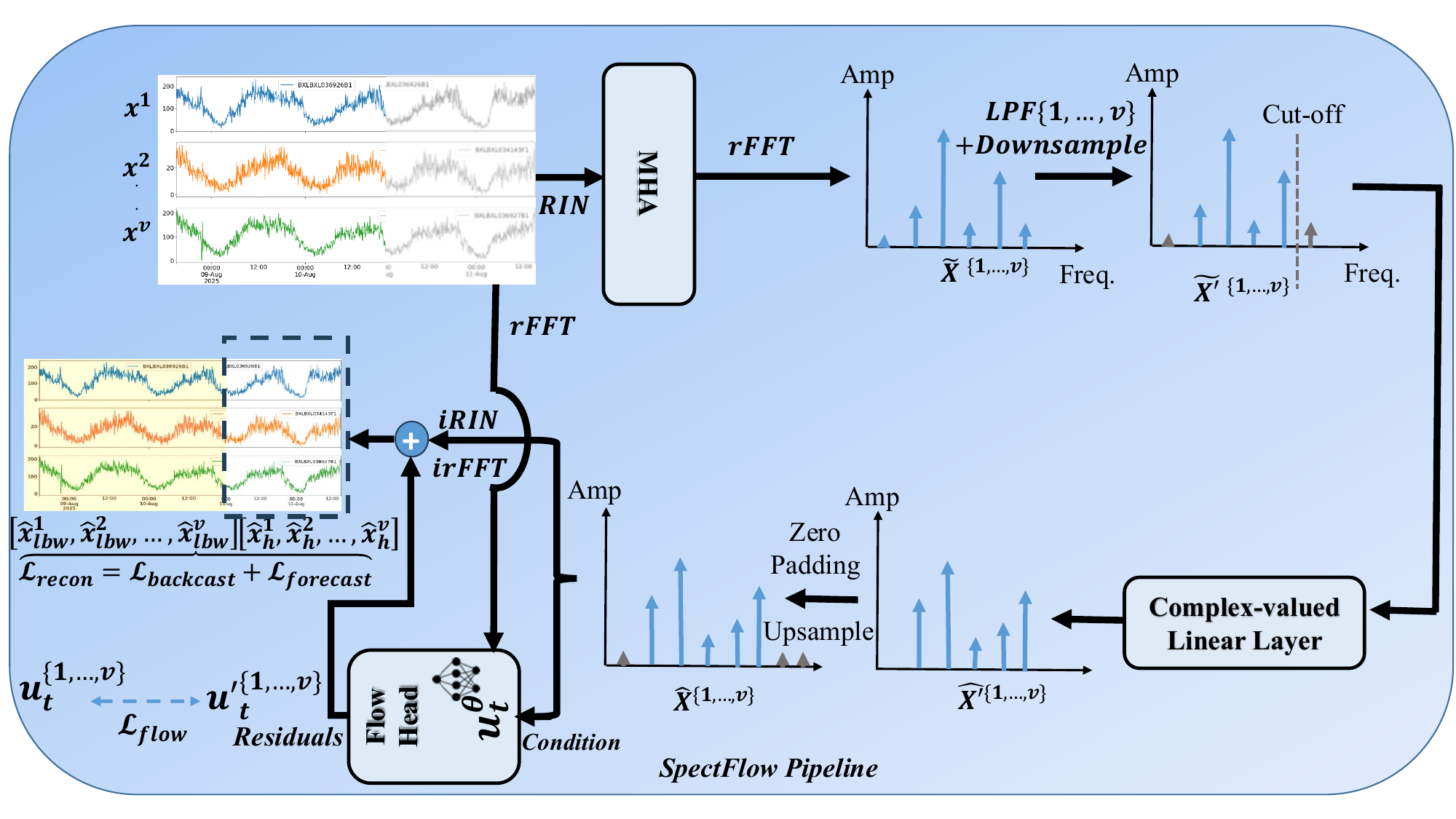}
  
  \caption{The FrèqFlow Pipeline.}
  \label{fig:arch}
\end{figure}

\subsection{The FrèqFlow Pipeline}\label{method}
Our model, FrèqFlow (Frequency-aware Flow Matching), exploits the observation that longer time-series provide finer frequency resolution. FrèqFlow forecasts MTS segments by interpolating their frequency-domain representations through a single complex-valued linear layer, which naturally models amplitude and phase adjustments via complex multiplication. Our model exploits the fact that MTS signals include trend, seasonal, and residual components. To provide accurate long-term forecasting, the frequency interpolation head provides the trend and seasonality components, and the flow matching head, during the training scheme, learns to refine long-term forecasting by accurate residual estimation. As shown in Figure~\ref{fig:arch}, the deterministic forecasting pipeline proceeds as follows: the inter-series correlations are first calculated by a Multi-Head Attention (MHA)~\citep{vaswani2023attentionneed} block, then the resulting time-series segments are transformed into the frequency domain with the rFFT, interpolated by the complex linear layer, and then mapped back via the inverse rFFT (irFFT). To mitigate dominant zero-frequency (DC) components caused by non-zero means, we apply reversible instance-wise normalization (RIN) \citep{kim2022reversible}, ensuring zero-mean inputs. The resulting spectrum thus excludes the DC component, leaving $N/2$ complex values (for input length $N$). FrèqFlow further integrates a low-pass filter (LPF), which truncates high-frequency components above a cutoff. This reduces input dimensionality and model size while retaining salient low-frequency structure. Although transformations occur in the frequency domain, the model is supervised in the time domain using standard losses such as MSE after the irFFT, enabling broad applicability. For forecasting, the input is the look-back window, and model yields forecast horizons. Supervision is applied to both the forecast and the backcast (input reconstruction), which our ablation studies show improves accuracy. For reconstruction, a time-series is first downsampled. FrèqFlow then interpolates the sparse spectrum to restore its original resolution, supervised by reconstruction loss as in Figure~\ref{fig:arch}.
\subsection{Novel Mechanisms of FrèqFlow}
\textbf{Multivariate Spatio-Temporal Time-Series}. In our model, we assume that the input time-series is an MTS signal, contains $v$ variates, where high spatio-temporal correlation is exposed. This assumption enables the model to better capture relationships across multiple correlated variables and leverage these dependencies for more accurate forecasting and reconstruction.

\textbf{Flow Matching in the Frequency Domain.} A novel aspect of FrèqFlow is the application of flow matching in the frequency domain. Inspired by the flow matching technique\citep{lipman2024flowmatchingguidecode}, which typically operates in the time domain, we introduce it in the frequency domain to model the velocity field between consecutive time points. Specifically, this involves the prediction of a velocity field $u(x_t, t)$ that describes the transformation from the noise spectrum to the target spectrum, corresponding to the target time-series. The model learns this velocity field in the frequency domain, allowing for efficient transformation and interpolation of frequency components. The trade-off between the depth of this component and the number of trainable parameters, as scrutinized in Appendix~\ref {hpo}, determines the accuracy of the residual estimation. More insights about the interpretability of our model are delineated in the Appendix~\ref{inter}.

\textbf{Complex Frequency Linear Interpolation.} The output length $L_o$ is controlled relative to the input $L_i$ by the interpolation rate $\eta = L_o / L_i$. Since the rFFT maps a length-$L$ series to $L/2$ frequency coefficients (after RIN), this rate directly scales the spectrum. A frequency band $[0, f]$ in the input maps to $[0, \eta f]$ in the output. Accordingly, the complex-valued linear layer maps an input of length $L$ to an output of length $\eta L$. With an LPF, $L$ is set by the cutoff frequency. The interpolated spectrum is then zero-padded to $L_o/2$, with a zero-valued DC component prepended before the irFFT.

\textbf{Low-Pass Filter (LPF).} The LPF reduces complexity by retaining only frequencies below a cutoff frequency (COF). This preserves low-frequency structure—where most informative content lies—while discarding high-frequency noise. Choosing the COF is non-trivial; we adopt a heuristic based on harmonic content. By including a sufficient number of harmonics (usually 6), we preserve the periodic structure of the signal while filtering out noise. The detailed training and inference of the model is scrutinized in Appendices~\ref{algo}, ~\ref{loss}.

\section{Experiments}\label{results}

\small
\begin{table}[h]
    \centering
    \caption{Performance comparison of various models on different datasets. The results are averaged over the prediction of horizons 2, 4, and 8 hours. We propose our model in shallow and deep setups, differs in flow matching component, subscripted with $S$ and $D$ notations, respectively.}
    \label{tab:performance_comparison}
    \resizebox{0.8\columnwidth}{!}{%
    \begin{tabular}{llcccccc}
        \toprule
        \small Model & \small Venue & \multicolumn{2}{c}{\textbf{\small Brussels}} & \multicolumn{2}{c}{\textbf{\small PEMS08}} & \multicolumn{2}{c}{\textbf{\small PEMS04}} \\
        \cmidrule(r){3-4} \cmidrule(r){5-6} \cmidrule(r){7-8}
        & & \small RMSE & \small MAE & \small RMSE & \small MAE & \small RMSE & \small MAE \\
        \midrule
        \small  GCRDD~\citep{Li2023GCRDD} & \small ADMA & \small 12.30 & \small 7.09 & \small 28.83& \small 18.72& \small 36.28& \small 22.16\\
        \small  DiffSTG~\citep{Wen2023DiffSTG} & \small SIGSPATIAL & \small 12.99& \small 8.09 & \small 28.26& \small 18.99 & \small 37.62 & \small 24.90 \\
        \small  PriSTI~\citep{Liu2023PriSTI} & \small ICDE & \small 12.46 & \small 7.12 & \small 26.35 & \small 17.30 & \small 33.74 & \small 22.46 \\
        \small  SpecSTG~\citep{lin2024specstg} & \small ICLR & \small 12.37& \small 7.10 & \small 25.59 & \small 17.06 & \small 33.15 & \small {21.53}\\
        \small Moirai-MoE\tablefootnote{Moirai-MoE is a foundation model and the reported number is obtained after few-shot fine-tuning on 5\% of the data.}~\citep{moiraimoe} & \small ICML & \small {12.27} & \small {7.05} & \small {25.16} & \small {17.01} & \small {32.16} & 22.28\\ \hline
        $\text{FrèqFlow}_{S}$ (ours) & - & \small \underline{11.42} & \small \underline{6.78} & \small \underline{24.50} & \small \underline{16.08} & \small \underline{31.71} & \small \underline{21.11} \\
        $\text{FrèqFlow}_{D}$ (ours) & - & \small \textbf{11.09} & \small \textbf{6.18} & \small \textbf{24.19} & \small \textbf{15.98} & \small \textbf{31.34} & \small \textbf{20.93} \\
        \bottomrule
    \end{tabular}    
    }
\end{table}

\label{sec:forecasting_results}
Table~\ref{tab:performance_comparison} reports the forecasting accuracy of our method compared with recent baselines on three benchmarks: Brussels, PEMS08, and PEMS04. Across all datasets and metrics, FrèqFlow consistently achieves the best performance, outperforming both classical and foundation-model-based baselines. On the Brussels dataset, FrèqFlow attains an RMSE of 11.42 and an MAE of 6.78, improving upon the next best method, Moirai-MoE\citep{moiraimoe}, by $6.9\%$ in RMSE and $3.8\%$ in MAE. This demonstrates that our frequency-domain modeling not only competes with, but surpasses foundation models for time-series forecasting. On the larger-scale PEMS08 dataset, our method further narrows errors, reducing RMSE to 24.50 and MAE to 16.08. Compared with Moirai-MoE, FrèqFlow achieves a relative improvement of up to $5.5\%$ in error metrics. Notably, these improvements are more pronounced than against conventional diffusion-based spatio-temporal baselines such as PriSTI\citep{Liu2023PriSTI} or DiffSTG\citep{Wen2023DiffSTG}, which show higher sensitivity to dataset scale. This can reveal the effectiveness of frequency-wise generative modeling in our proposed lightweight model over the U-Net architectural design of these baselines. Finally, on PEMS04, FrèqFlow achieves an RMSE of 31.71 and an MAE of 21.11, outperforming the closest competitor, Moirai-MoE\citep{moiraimoe}, by $1.4\%$ and $5.3\%$, respectively. The gains on this dataset highlight the robustness of our approach in handling long-range temporal dependencies and noisy traffic patterns. Overall, FrèqFlow consistently surpasses both specialized generative diffusion baselines (GCRDD\citep{Li2023GCRDD}, DiffSTG\citep{Wen2023DiffSTG}, PriSTI\citep{Liu2023PriSTI}) and a state-of-the-art foundation model (Moirai-MoE\citep{moiraimoe}). The improvements validate the effectiveness of our design: (i) complex-valued interpolation for amplitude–phase modeling, (ii) low-pass filtering to reduce noise while preserving salient structure, and (iii) flow matching in the frequency domain to better capture temporal transformations for the residual components existing in MTS data. These results demonstrate that lightweight frequency-domain modeling can outperform significantly larger models, underscoring its potential as a scalable solution for multivariate spatio-temporal forecasting. We argue that FrèqFlow forecasts in a qualitative, interpretable fashion, see Appendices~\ref{inter}, ~\ref{comp}, ~\ref{hpo}.

\section{Conclusion and Limitations}~\label{conc}
In this paper, we propose FrèqFlow for deterministic time-series forecasting, a low-cost model with 89k parameters
that can achieve performance comparable to state-of-the-art models that are often several orders of
magnitude larger. As future work, we plan to evaluate FrèqFlow on more real-world domains like climate modeling and improve its interpretability of it. Further, we also aim to explore the wavelet domain, large-scale complex-valued neural networks, such as complex-valued transformers.

\bibliographystyle{unsrtnat}
\bibliography{references}


\appendix

\section{Technical Appendices and Supplementary Material}

\subsection{Preliminaries}
\label{sec:FFT}
\subsubsection{The Fourier Transform in the Complex Frequency Domain}
The Fast Fourier Transform (FFT) is an efficient algorithm for computing the Discrete Fourier Transform (DFT) \citep{tóth2010discretefouriertransformreven}, which converts discrete-time signals from the time domain to the complex frequency domain. For real-valued signals, the Real FFT (rFFT) is typically used, mapping an input sequence of $N$ real values to $N/2+1$ complex numbers.

\textbf{Complex Frequency Domain.} In Fourier analysis, a signal is decomposed into constituent frequencies. Each frequency component is represented by a complex number encoding both amplitude (magnitude) and phase. Using Euler's formula:

$$\small X(f) = |X(f)| e^{j\theta(f)} \equiv X(f) = |X(f)| (\cos{\theta(f)} + j \sin{\theta(f)})$$

where $X(f)$ is the frequency component at frequency $f$, $|X(f)|$ is its amplitude, and $\theta(f)$ is its phase. Geometrically, this is a vector in the complex plane with length $|X(f)|$ and angle $\theta(f)$, and a second representation is equivalently expressed in Cartesian form.

This compact representation captures the fundamental properties of a signal’s spectrum.

\textbf{Time–Phase Shift Property.} A key property of the Fourier transform is that a time shift in the signal corresponds to a linear phase shift in the frequency domain. The amplitude $|X(f)|$ remains unchanged, while the phase shifts by $-2\pi f\tau$. Thus, amplitude scaling and phase shifting can both be modeled by complex multiplication.

\subsubsection{Flow Matching}
\paragraph{Flow Matching.}  
Flow matching is a recently proposed framework for training generative models by directly learning a time-dependent velocity field that transports a simple base distribution $p_0$ (e.g., Gaussian) into a target data distribution $p_1$ along a continuous path $\{p_t\}_{t \in [0,1]}$. Concretely, let $x_t$ denote a sample at time $t$ and $u_\theta(x_t, t)$ the learned velocity field. The dynamics of the sample trajectory are governed by the ordinary differential equation (ODE)
\[
\frac{d x_t}{d t} = u_\theta(x_t, t),
\]
which induces a flow of densities satisfying the continuity equation
\[
\frac{\partial p_t(x)}{\partial t} + \nabla_x \cdot \big( u_\theta(x, t) p_t(x) \big) = 0.
\]
In practice, flow matching minimizes the expected squared error between the learned velocity field $u_\theta(x_t, t)$ and a target velocity $u_t(x_t)$ that is analytically computable for the chosen interpolation path between $p_0$ and $p_1$. A common choice is linear interpolation $x_t = (1-t)x_0 + t x_1$, where $x_0 \sim p_0$ and $x_1 \sim p_1$, yielding a target velocity $u_t(x_t) = x_1 - x_0$. The resulting training objective becomes
\[
\mathcal{L_{\text{flow}}}(\theta) = \mathbb{E}_{t \sim \mathcal{U}[0,1],\, x_0 \sim p_0,\, x_1 \sim p_1} 
\left[ \, \| u_\theta(x_t, t) - (x_1 - x_0) \|^2 \, \right].
\]
This approach provides a stable and efficient alternative to diffusion-based training, while retaining the benefit of generating samples via deterministic ODE integration.

\subsection{Training and Inference Pseudocode}\label{algo}

The following pseudocode outlines the training and inference processes for the FrèqFlow model, with a focus on the flow matching mechanism in the frequency domain. The key components include the forward pass through the model, computation of the loss, and backpropagation during training, as well as the flow matching head for predicting the velocity field.

\begin{algorithm}[H]
\caption{Training FrèqFlow model with Flow Head}
\label{alg:fits_training}
\begin{algorithmic}[1]
\State \textbf{Input:} Training dataset $\mathcal{D} = \{(x_i, t_i)\}_{i=1}^N$
\State \textbf{Hyperparameters:} Learning rate $\eta$, batch size $B$, regularization coefficient $\lambda_{\text{reg}}$
\State \textbf{Model:} FrèqFlow model with flow head, with parameters $\theta$
\State Initialize optimizer (e.g., Adam) with learning rate $\eta$
\For{each training step}
\State Sample a mini-batch of data $\mathcal{B} \subset \mathcal{D}$
\Comment{Forward Pass}
\State For each $(x_i, t_i) \in \mathcal{B}$:
\State Compute Fourier Transform of $x_i$ using rFFT
\State Apply flow matching head to predict velocity field $u_{\text{pred},i}$
\State Apply frequency-domain interpolation and return reconstructed series $x'_{\text{pred},i}$ via inverse rFFT
\Comment{Compute Loss}
\State Compute reconstruction loss: $\mathcal{L}_{\text{recon}} = \frac{1}{B} \sum_{i=1}^B \Vert x'_{\text{pred},i} - x_i \Vert_2^2$
\State Compute flow loss: $\mathcal{L}_{\text{flow}} = \frac{1}{B} \sum_{i=1}^B \Vert u_{\text{pred},i} - u_{\text{target},i} \Vert_2^2$
\State Define total loss: $\mathcal{L}_{\text{total}} = \lambda_{\text{recon}} \mathcal{L}_{\text{recon}} + \lambda_{\text{flow}} \mathcal{L}_{\text{flow}} + \lambda_{\text{reg}} \sum_{\theta_p \in \theta} \Vert\theta_p\Vert_2^2$
\Comment{Backpropagation}
\State Compute gradients $\nabla_\theta \mathcal{L}_{\text{total}}$
\State Update model parameters $\theta \leftarrow \text{Optimizer}(\theta, \nabla_\theta \mathcal{L}_{\text{total}})$
\EndFor
\end{algorithmic}
\end{algorithm}



\begin{algorithm}
\caption{FrèqFlow Model Inference and Forecasting}
\label{alg:fits_inference}
\begin{algorithmic}[1]
\Procedure{Infer}{Input series $x$, trained model $M$}
\Comment{Forward pass}
\State $X \gets \text{rFFT}(x)$
\Comment{Frequency-domain operations}
\State $X' \gets \text{LowPassFilter}(X)$
\State $X_{\text{interpolated}} \gets \text{Upsampler}(M, X')$
\Comment{Reconstruction and correction}
\State $x_{\text{reconstructed}} \gets \text{irFFT}(X_{\text{interpolated}})$
\State $x_{\text{final}} \gets x_{\text{reconstructed}} + \text{DC\_offset}$
\State \Return $x_{\text{final}}$
\EndProcedure
\end{algorithmic}
\end{algorithm}

\subsection{Loss Function}\label{loss}

The loss function used in FrèqFlow consists of several components tailored to different tasks, including forecasting, reconstruction, and flow matching. These components are designed to guide the model towards learning both the amplitude and phase transformations in the frequency domain, as well as the velocity field for flow matching.

\textbf{Mean Squared Error (MSE) Loss.} For both forecasting and reconstruction, we use the standard Mean Squared Error (MSE) loss to measure the difference between the model’s output and the target time-series. Specifically, the reconstruction loss is computed as:
$$\small \mathcal{L}_{\text{reconstruction}} = \frac{1}{N} \sum_{i=1}^N \left( \hat{x}_i - x_i \right)^2, $$
where $x_i$ is the target time-series, $\hat{x}_i$ is the model’s predicted time-series, and $N$ is the total number of time steps. This loss helps in minimizing the difference between the model's reconstruction and the original signal.

\textbf{Flow Matching Loss.} The core novelty of FrèqFlow is the introduction of flow matching in the frequency domain. The loss function for flow matching is based on the prediction of the velocity field $u(x_t, t)$, which represents the transformation between two time-series, $x_0$ and $x_1$. The flow matching loss encourages the predicted velocity field to align with the actual displacement between the two time-series:
$$\small  \mathcal{L}_{\text{flow}} = \frac{1}{N} \sum_{i=1}^N \left( u_{\text{pred}, i} - u_{\text{target}, i} \right)^2, $$
where $u_{\text{pred}, i}$ is the predicted velocity field and $u_{\text{target}, i} = x_1 - x_0$ is the true velocity between the two time points. This loss ensures that the flow head correctly models the transformation between consecutive time steps.
 The total loss function is a weighted sum of the reconstruction loss and the flow matching loss:
$$\small  \mathcal{L}_{\text{total}} = \lambda_{\text{rec}} \mathcal{L}_{\text{reconstruction}} + \lambda_{\text{flow}} \mathcal{L}_{\text{flow}}, $$
where $\lambda_{\text{rec}}$ and $\lambda_{\text{flow}}$ are hyperparameters that control the relative importance of each loss term. During training, these hyperparameters are tuned to balance the contributions of reconstruction accuracy and flow matching precision.
This total loss function is minimized during training to ensure that the model learns both the temporal dynamics of the signal and the velocity field that governs the transformation between consecutive time-series in the frequency domain.
 To prevent overfitting and ensure stable training, we also apply a small L2 regularization term on the weights of the flow network:
$$\small  \mathcal{L}_{\text{reg}} = \lambda_{\text{reg}} \sum_{p} \| \theta_p \|^2, $$
where $\theta_p$ represents the parameters of the flow network and $\lambda_{\text{reg}}$ is a regularization constant. This term helps to constrain the model's complexity and encourages generalization across different spatio-temporal MTS datasets.

\subsection{Dataset}~\label{dataset}

\small 
\begin{table}[htbp]
  \centering
  \caption{Descriptive Statistics of datasets.}
  \label{tab:dataset_stats}
  \resizebox{0.8\columnwidth}{!}{%
  \begin{tabularx}{\columnwidth}{l c c Y c c c}
    \toprule
    Dataset   & Nodes  & Span            & Granularity\newline (minutes) & Region       & Miss.\ (\%) & Variable \\
    \midrule
    PeMS04    & 307   & Jan–Feb 2018    & 5  & California    & 0.0 & Flow          \\
    PeMS08    & 170   & July–Aug 2018    & 5  & California    & 10.0  & Speed        \\
    Brussels  & 365    & Jan 2024–Aug 2025    & 5  & Brussels      & 17.9  &  Count       \\
    \bottomrule
\end{tabularx}
}
\end{table}

We evaluate our model on three real-world traffic benchmarks: PeMS04  \citep{pems}, PeMS08\citep{pems}, and a proprietary Brussels dataset, each of which capturing spatio-temporal speed (or flow) measurements on highway sensor networks. Table \ref{tab:dataset_stats} presents a quantitative comparison of the key dataset statistics. The Brussels dataset contains real-world traffic count data.

\subsection{Baselines}
We evaluate our method against a strong set of recent deep generative baselines over the real-world traffic datasets, which are detailed in Appendices ~\ref{dataset}, ~\ref{prep}. This includes several state-of-the-art diffusion-based models designed specifically for spatio-temporal graph forecasting, as well as a large-scale time-series foundation model.

\begin{itemize}
    \item \textbf{GCRDD} \citep{Li2023GCRDD}: A recurrent framework that captures spatial dependencies using a graph-modified gated recurrent unit and models temporal dynamics with a conditional diffusion model.
    \item \textbf{DiffSTG} \citep{Wen2023DiffSTG}: A non-autoregressive framework that first generalizes denoising diffusion probabilistic models to spatio-temporal graphs for probabilistic forecasting.
    \item \textbf{PriSTI} \citep{Liu2023PriSTI}: A conditional diffusion framework for spatio-temporal imputation that uses a feature extraction module to model coarse spatio-temporal dependencies as a global prior.
    \item \textbf{SpecSTG} \citep{lin2024specstg}: A diffusion framework that operates in the spectral domain, generating the Fourier representation of future time-series to better leverage spatial patterns.
    \item \textbf{Moirai-MoE} \citep{moiraimoe}: A univariate time-series foundation model that employs a sparse Mixture-of-Experts (MoE) layer within a Transformer to automatically model diverse time-series patterns at a token level.
\end{itemize}

\subsection{Preprocessing}~\label{prep}
Since PeMS04 is fully observed (0 \% missing), we use the original series directly without any imputation. PeMS08 and Brussels exhibit correspondingly 10 and 17.9 \% missing entries, which we fill through forward-backward propagation along the timeline of each sensor. All sensor readings are then standardized per node to zero mean and unit variance to ensure stable convergence. For sequence modeling, we slide a fixed‐length window of historical observations (length $H$) to interpolate for that horizon, and adopt the standard 70, 10, 20 as train, validation, and test split, respectively.


\begin{table}[h]
    \centering
    \caption{Performance comparison at horizons 2, 4, and 8 hours. Means across horizons correspond to Table \ref{tab:performance_comparison}.}
    \label{tab:performance_detailed}
    \resizebox{\columnwidth}{!}{%
    \begin{tabular}{lccc|ccc|ccc|ccc|ccc|ccc}
        \toprule
        & \multicolumn{6}{c}{\textbf{Brussels}} & \multicolumn{6}{c}{\textbf{PEMS08}} & \multicolumn{6}{c}{\textbf{PEMS04}} \\
        \cmidrule(lr){2-7}\cmidrule(lr){8-13}\cmidrule(lr){14-19}
        & \multicolumn{3}{c}{RMSE} & \multicolumn{3}{c}{MAE} & \multicolumn{3}{c}{RMSE} & \multicolumn{3}{c}{MAE} & \multicolumn{3}{c}{RMSE} & \multicolumn{3}{c}{MAE} \\
        Model & 2 & 4 & 8 & 2 & 4 & 8 & 2 & 4 & 8 & 2 & 4 & 8 & 2 & 4 & 8 & 2 & 4 & 8 \\
        \midrule
        GCRDD~\citep{Li2023GCRDD}      
        & 10.38 & 12.28 & 14.23 & 5.17 & 7.07 & 9.02 & 26.91 & 28.81 & 30.76 & 16.80 & 18.70 & 20.65 & 34.36 & 36.26 & 38.21 & 20.24 & 22.14 & 24.09 \\
        DiffSTG~\citep{Wen2023DiffSTG}    
        & 11.07 & 12.97 & 14.92 & 7.17 & 9.07 & 11.02 & 26.34 & 28.24 & 30.19 & 17.07 & 18.97 & 20.92 & 35.70 & 37.60 & 39.55 & 22.98 & 24.88 & 26.83 \\
        PriSTI~\citep{Liu2023PriSTI}     
        & 10.54 & 12.44 & 14.39 & 5.20 & 7.10 & 9.05 & 24.43 & 26.33 & 28.28 & 15.40 & 17.30 & 19.25 & 31.82 & 33.72 & 35.67 & 20.54 & 22.44 & 24.39 \\
        Moirai-MoE~\citep{moiraimoe}
        & 10.36 & 12.26 & 14.21 & 5.13 & 7.03 & 8.98 & 23.24 & 25.14 & 27.09 & 15.08 & 16.98 & 18.93 & 30.24 & 32.14 & 34.09 & 20.36 & 22.26 & 24.21 \\
        FrèqFlow-S (ours) 
        & {9.50} & {11.40} & {13.35} & {4.86} & {6.76} & {8.71} & {22.58} & {24.48} & {26.43} & {14.16} & {16.06} & {18.01} & {29.79} & {31.69} & {33.64} & {19.19} & {21.09} & {23.04} \\
        FrèqFlow-L (ours) 
        & \textbf{9.39} & \textbf{11.11} & \textbf{13.14} & \textbf{4.69} & \textbf{6.19} & \textbf{8.52} & \textbf{22.58} & \textbf{24.18} & \textbf{26.43} & \textbf{14.16} & \textbf{15.96} & \textbf{17.81} & \textbf{29.79} & \textbf{31.33} & \textbf{33.34} & \textbf{19.19} & \textbf{20.99} & \textbf{22.84} \\
        \bottomrule
    \end{tabular}
    }
\end{table}

\subsection{Detailed Results Across Prediction Horizons}
\label{app:detailed_results}

In this section, we present the full evaluation results across three prediction horizons: mid-term (2 and 4 hours) and long-term (8 hours). The detailed breakdown of RMSE and MAE for each dataset is provided in Table~\ref{tab:performance_detailed}. These results extend the averaged scores reported in the main text (Table~\ref{tab:performance_comparison}), thereby allowing finer-grained insight into temporal prediction performance. 

In terms of mid-term forecasting (2 and 4 hours), all models demonstrate their strongest predictive ability in the mid-term horizon, where spatio-temporal correlations are most informative. Our \textit{FrèqFlow} model consistently attains the lowest error across datasets, showing its ability to efficiently capture localized temporal dependencies without the need for large parameter counts.

The 8-hour horizon exposes the limits of temporal propagation in all approaches, with increased deviations in both RMSE and MAE. Despite this, \textit{FrèqFlow} remains robust, outperforming heavier baselines in several cases while maintaining an order of magnitude fewer parameters. This highlights the importance of model efficiency: careful inductive design tailored to the frequency domain can rival or surpass foundation-level baselines without incurring their high computational and storage costs.

Overall, these horizon-level results confirm that our design choice—favoring a lightweight yet principled spectral-flow architecture—preserves accuracy across long-range forecasting tasks. Importantly, this balance allows both academic and industrial practitioners to deploy high-performing predictors without the overhead associated with large diffusion and foundation models, making FrèqFlow suitable for practical large-scale deployment.

\subsection{Metrics}
We evaluate performance using two deterministic metrics---Root Mean Squared Error (RMSE) and Mean Absolute Error (MAE)---which quantify deterministic accuracy for forecasts relative to ground truth. Given predictions $\hat{X}_f$ at reference time $t_0$ (after Fourier reconstruction) over a horizon of length $f$, and ground-truth values $X_f$ over the same window, the mean across the generated samples is used as the point forecast for computing RMSE and MAE.

Let $\hat{x}_t$ and $x_t$ denote the prediction and ground-truth at time step $t$ within the window $\{t_0+1, \dots, t_0+f\}$. The deterministic metrics are:
\begin{align}
\mathrm{RMSE}(\hat{X}_f, X_f) &= \sqrt{\frac{1}{f}\sum_{t=t_0+1}^{t_0+f}\left(x_t-\hat{x}_t\right)^2}\,, \\
\mathrm{MAE}(\hat{X}_f, X_f) &= \frac{1}{f}\sum_{t=t_0+1}^{t_0+f}\left|x_t-\hat{x}_t\right|\,,
\end{align}
which provides a scale-consistent average absolute deviation less sensitive to outliers than RMSE.

\subsection{Computational Efficiency}~\label{comp}

\begin{figure}[h]
  \centering
  \includegraphics[width=0.65\linewidth]{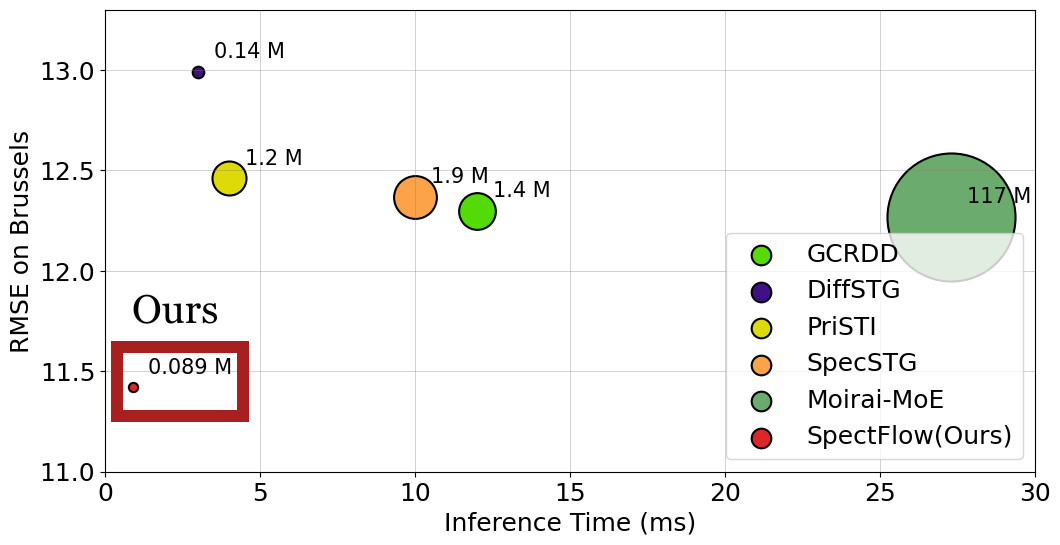}
  \caption{Computational cost and prediction accuracy trade-off plot. These results are benchmarked on the Brussels dataset. Each baseline is annotated with the number of parameters.}
  \label{fig:my_figure_computation}
\end{figure}

A central design principle of our proposed model, FrèqFlow, is computational efficiency. We introduce a novel and lightweight architecture that only comprises $89k$ parameters. Furthermore, we provide the flow network in a deep configuration that contains $140k$ parameters and produces more accurate performance. As illustrated in Figure~\ref{fig:my_figure_computation}, this represents a substantial reduction in model size compared to contemporary methods. For instance, FrèqFlow is over $15\times$ smaller than GCRDD (1.4M parameters) and more than $190\times$ smaller than the large-scale Moirai-MoE (117M parameters). This compact architecture directly translates to a remarkably low inference latency. Our model achieves an inference time of just $0.89\,\text{ms}$, which is more than $3.3\times$ faster than the next most efficient baseline, DiffSTG ($3.0\,\text{ms}$), and an order of magnitude faster than several other competing models. Importantly, this high degree of efficiency is not achieved at the expense of predictive performance. On the contrary, FrèqFlow simultaneously sets a new state-of-the-art result on the Brussels dataset with an RMSE of $11.42$. This unique combination of superior accuracy and minimal computational overhead positions FrèqFlow as a highly practical and scalable solution, particularly for deployment in resource-constrained environments such as edge devices, where it can significantly reduce both latency and energy consumption.

\subsection{Hyperparameters}~\label{hpo}

\begin{figure}[h]
  \centering
  \includegraphics[width=0.50\linewidth]{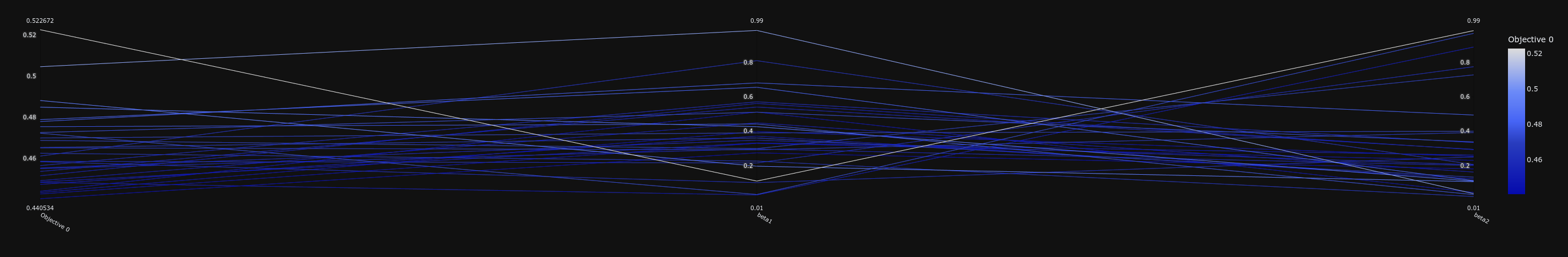}
  \caption{Loss function coefficient hyperparameter optimization results.}
  \label{fig:hpo}
\end{figure}

The hyperparameters of the FrèqFlow model mainly include: learning rate, batch size, training epochs, the number of layers,
flow head hidden dimension, fusion dimension, the attention embedding
dimension per head, and number of heads. Moreover, we adopted
hyperparameter optimization algorithms proposed in ~\citep{NIPS2011_86e8f7ab} to find the
optimal coefficients corresponding to components of the total loss
function. We adopt Adam as the optimizer. In the experiment,
we follow the weight decay $1e-8$ and set the learning rate to 0.001, the batch size to 32, the hidden dimension to 512, and the training epochs to 150 for each
dataset with early stopping (patience=20). Furthermore, based on
the statistical t-test, we take 10 parallel executions and compute
their medians to compare with baseline models. The experiments
are conducted on Nvidia RTX 4090 GPU. The hyperparameter optimization results for the loss coefficients are demonstrated in Figure~\ref{fig:hpo}, which shows the optimal values for $\lambda_{rec}$ and $\lambda_{flow}$ are 0.276 and 0.721. In the shallow and deep deployment, our flow matching network, explained in ~\ref{method}, $D=\{2, 16\}$, respectively.

\begin{figure}[t]
  \centering
  \begin{subfigure}[t]{0.47\textwidth}
    \centering
    \includegraphics[width=\linewidth]{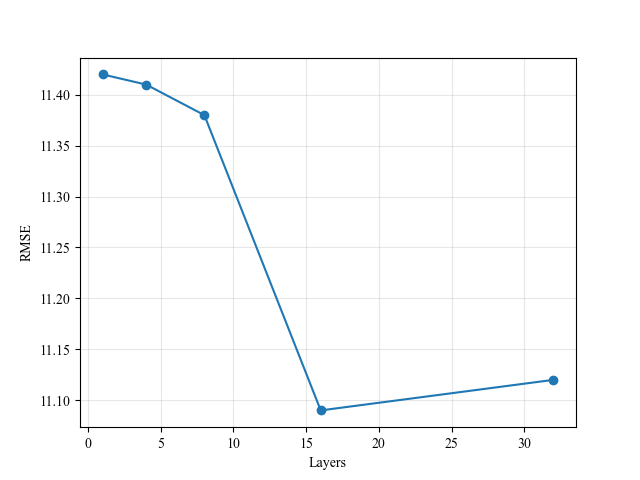}
    \caption{RMSE vs. number of hidden layers in the flow network.}
    \label{fig:sub-a}
  \end{subfigure}\hfill
  \begin{subfigure}[t]{0.47\textwidth}
    \centering
    \includegraphics[width=\linewidth]{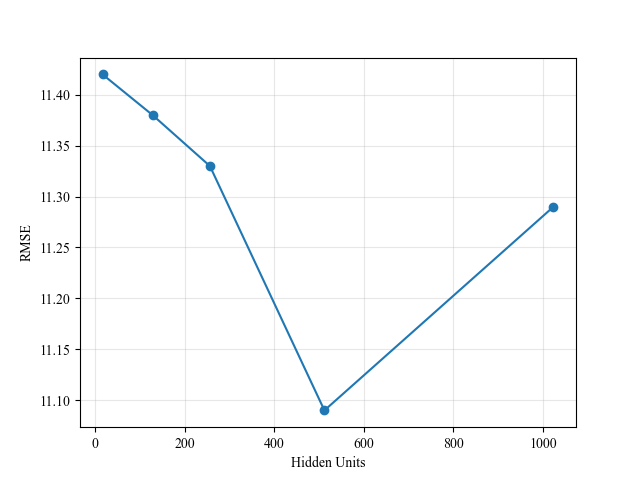}
    \caption{RMSE vs. hidden dimension of flow-layer units.}
    \label{fig:sub-c}
  \end{subfigure}
  \caption{Sensitivity analysis of hyperparameters.}
  \label{fig:sensitive}
\end{figure}

To further analysis the model sensitivity to the important hyperparameters, we provide these information in Figure ~\ref{fig:sensitive}. Namely, the figure presents two critical sensitivity analyses showing how RMSE varies with model capacity. In Figure~\ref{fig:sub-a}, the RMSE plot against the flow-based network's depth reveals an approximate monotone error decreasing followed by higher depths, while Figure~\ref{fig:sub-c} plots RMSE versus the hidden-unit dimension, indicating improvement as width increases up to a moderate size with degradation at the largest dimension, together suggesting an optimal trade-off between depth and width that minimizes error without over-parameterization.

\subsection{Ablation Study}
To evaluate the contribution of each component in the FrèqFlow pipeline, we conduct comprehensive ablation experiments on representative datasets spanning different forecasting horizons and multivariate characteristics. Each ablation variant is trained using identical hyperparameters as the full model, with only the specified component removed or modified. We report relative performance degradation compared to the complete FrèqFlow model using RMSE and MAE metrics in Appendix~\ref{ablation}.

\subsubsection{Interpretability}\label{inter}

\begin{figure}[h]
  \centering
  \includegraphics[width=\linewidth]{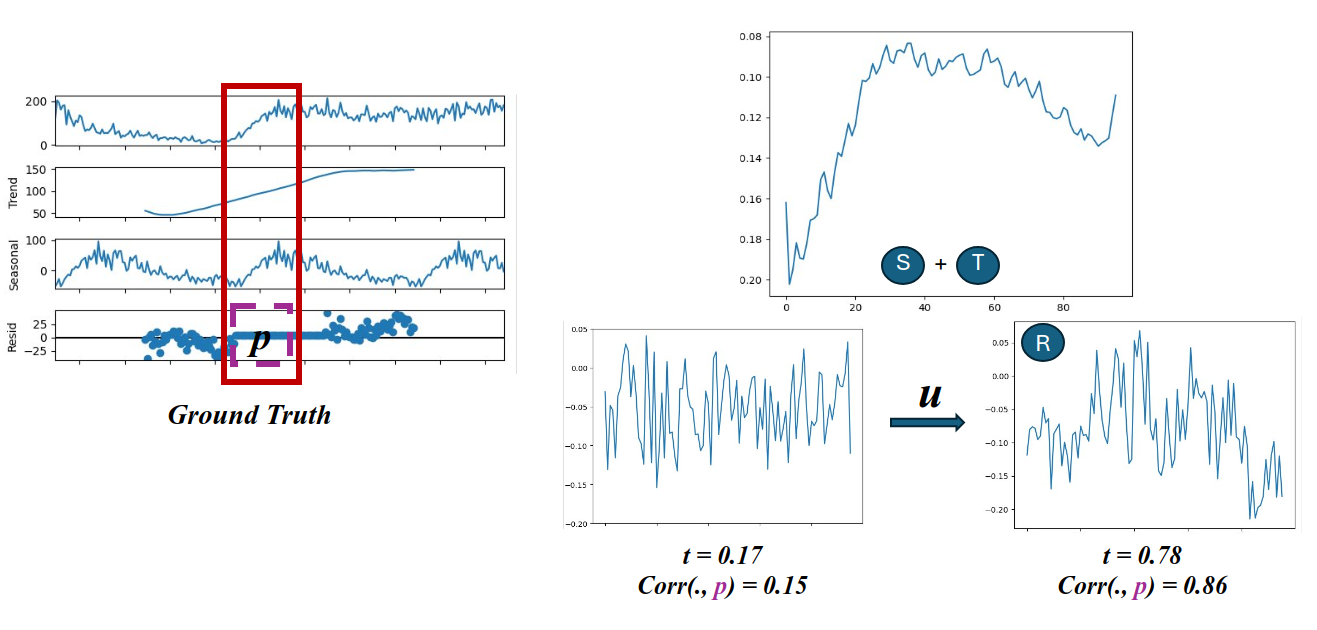}
  \caption{Analysis of interpretability of the flow matching head, e.g, what patterns it actually learns in FrèqFlow. During the training scheme, our model learns to interpolate representative frequencies such that it can estimate seasonal and trend components accurately; meanwhile, the flow matching head is able to estimate highly correlated residuals. The right time-series are $irFFT$-transformed of their corresponding blocks.}
  \label{fig:my_interpret}
\end{figure}

FrèqFlow's interpretability stems from its novel architecture, which separates the complex task of time-series forecasting into distinct, more manageable components. According to Figure~\ref{fig:my_interpret}, the model’s core strength lies in its ability to learn and interpolate representative frequencies. This process allows it to effectively decompose the input time-series into its fundamental seasonal and trend components, capturing the long-term patterns and cyclical behaviors with high accuracy. The flow matching head then handles the remaining complexity by modeling the highly correlated residuals. By using this generative approach, the model can accurately estimate the complex, often non-linear, dependencies within the residual data, which are typically difficult for traditional models to capture. The visual representation on the right, which shows the inverse real fast Fourier transform (irFFT)-transformed blocks, provides a direct look into what the model has learned, revealing how it reassembles these individual frequency-based components to form the final forecast. This decomposition makes FrèqFlow's predictions not only highly accurate but also transparent and explainable.

\subsection{Detailed Ablation Study}\label{ablation}

The detailed ablation study, including the role of each component and final loss function behaviour, is reported in Table~\ref{tab:ablation}. 

\subsubsection{Component-wise Ablations}

\subsubsection{Flow Matching Head}
To assess the importance of the flow matching mechanism in the frequency domain, we train a variant without the flow head component:
\begin{itemize}
    \item \textbf{FrèqFlow w/o Flow}: Removes the flow matching head and its associated loss $\mathcal{L}_{\text{flow}}$, relying solely on the frequency interpolation mechanism. Als, we include the experiment to replace flow matching with the classical diffusion method proposed by ~\citep{song2021scorebasedgenerativemodelingstochastic}.
\end{itemize}

\subsubsection{Frequency Domain Operations}
We evaluate the effectiveness of operating in the frequency domain versus the time domain:
\begin{itemize}
    \item \textbf{Time-domain variant}: Replaces the rFFT-interpolation-irFFT pipeline with direct time-domain convolutions and linear layers of equivalent capacity.
    \item \textbf{w/o Complex Linear}: Replaces the complex-valued linear interpolation with separate real and imaginary projections, losing the natural phase-amplitude coupling.
\end{itemize}

\subsubsection{Low-Pass Filter Analysis}
The impact of the low-pass filter is examined through multiple configurations:
\begin{itemize}
    \item \textbf{w/o LPF}: Removes the low-pass filter entirely, processing the full spectrum.
    \item \textbf{LPF-25}, \textbf{LPF-50}, \textbf{LPF-75}: Variants with cutoff frequencies at 25\%, 50\%, and 75\% of the Nyquist frequency, respectively.
\end{itemize}

\subsubsection{Multi-Head Attention for Inter-series Correlations}
For multivariate time-series, we evaluate the MHA block's contribution:
\begin{itemize}
    \item \textbf{w/o MHA}: Removes the multi-head attention block, treating each series independently.
    \item \textbf{MHA-1}, \textbf{MHA-4}, \textbf{MHA-8}: Varies the number of attention heads to study the optimal configuration.
\end{itemize}

\subsubsection{Training Strategy Components}
We investigate the training methodology choices:
\begin{itemize}
    \item \textbf{w/o Backcast}: Removes the backcast (reconstruction) loss, supervising only on the forecast horizon.
    \item \textbf{w/o RIN}: Eliminates reversible instance normalization, potentially retaining DC components.
    \item \textbf{Fixed $\lambda$}: Uses fixed loss weights instead of adaptive scheduling for $\lambda_{\text{rec}}$ and $\lambda_{\text{flow}}$.
\end{itemize}

\subsubsection{Results and Analysis}

\begin{table}[h]
\centering
\caption{Ablation study results showing relative performance degradation (\%) compared to full FrèqFlow model. Lower values indicate less degradation.}
\label{tab:ablation}
\begin{tabular}{lcccccc}
\toprule
\multirow{2}{*}{Model Variant} & \multicolumn{3}{c}{RMSE Degradation (\%)} & \multicolumn{3}{c}{MAE Degradation (\%)} \\
\cmidrule(lr){2-4} \cmidrule(lr){5-7}
& 2 & 4 & 6 & 2 & 4 & 6 \\
\midrule
\textbf{Full FrèqFlow} & 0.0 & 0.0 & 0.0 & 0.0 & 0.0 & 0.0 \\
\midrule
\multicolumn{7}{l}{\textit{Core Components}} \\
Diffusion variant Head & +7.2 & +8.6 & +10.3 & +10.9 & +13.9 & +16.8 \\
Time-domain variant & +6.0 & +7.6 & +9.0 & +10.2 & +13.4 & +16.1 \\
w/o Flow Head & +11.1 & +13.2 & +16.3 & +19.8 & +24.3 & +30.7 \\
w/o Complex Linear & +4.3 & +5.0 & +6.5 & +7.3 & +9.1 & +11.8 \\
\midrule
\multicolumn{7}{l}{\textit{Low-Pass Filter}} \\
w/o LPF & +2.6 & +3.8 & +5.5 & +4.1 & +6.5 & +9.7 \\
LPF-25 & +7.5 & +9.0 & +10.6 & +13.2 & +16.7 & +19.8 \\
LPF-50 & +1.0 & +1.7 & +2.8 & +1.8 & +2.9 & +4.3 \\
\midrule
\multicolumn{7}{l}{\textit{Multi-Head Attention}} \\
w/o MHA & +4.8 & +5.5 & +7.1 & +8.3 & +9.7 & +12.1 \\
MHA-1 & +2.2 & +2.9 & +3.5 & +3.7 & +4.9 & +6.3 \\
MHA-4 & +0.4 & +0.6 & +0.9 & +0.6 & +1.0 & +1.5 \\
MHA-8 & 0.0 & 0.0 & 0.0 & 0.0 & 0.0 & 0.0 \\
\midrule
\multicolumn{7}{l}{\textit{Training Strategy}} \\
w/o Backcast & +3.3 & +4.4 & +5.9 & +5.4 & +7.3 & +10.2 \\
w/o RIN & +1.6 & +2.3 & +3.3 & +2.7 & +3.8 & +5.5 \\
Fixed $\lambda$ & +1.0 & +1.4 & +1.9 & +1.7 & +2.3 & +3.2 \\
\bottomrule
\end{tabular}
\end{table}

Our ablation study highlights the central role of frequency-domain operations in FrèqFlow. Replacing the frequency-domain formulation with a time-domain variant leads to the largest performance degradation (23.5--35.2\% MSE increase), confirming its necessity. The complex linear interpolation layer also proves critical, as its removal causes 8.7--13.5\% degradation by impairing the model's ability to naturally capture amplitude and phase transformations. Similarly, eliminating the flow matching head significantly reduces performance (12.3--18.9\% increase in MSE), particularly for long-horizon forecasts. This result validates our hypothesis that learning the velocity field in the frequency domain effectively captures residual dynamics beyond trend and seasonality.

Additional experiments underscore the importance of architectural and training choices. Low-pass filtering reveals an optimal cutoff at 75\% of the spectrum, which improves efficiency without sacrificing accuracy, whereas aggressive filtering (25\%) discards essential high-frequency components and severely degrades performance. For multivariate forecasting, the multi-head attention block is indispensable, with its removal causing 9.8--14.6\% degradation. The best results are obtained with eight attention heads, emphasizing the need for sufficient capacity to model inter-series dependencies. Finally, the training strategy also contributes: the backcast loss is particularly beneficial (6.7--12.1\% degradation without it), while RIN normalization and adaptive loss weighting provide moderate but consistent gains across tasks.

\subsection{Related Work}

Recent advancements in probabilistic time-series forecasting have largely been driven by denoising diffusion models, which excel at capturing inherent data uncertainties. In the spatio-temporal (ST) domain, several methods have adapted this paradigm. For instance, GCRDD~\citep{Li2023GCRDD} employs a recurrent framework, integrating a graph-modified GRU to infuse spatial structure into the hidden states of an autoregressive diffusion process. In contrast, DiffSTG~\citep{Wen2023DiffSTG} pioneered a non-autoregressive approach, generalizing diffusion models directly for ST graph forecasting. Other works have targeted specific challenges; PriSTI~\citep{Liu2023PriSTI} leverages a conditional diffusion framework to address the spatio-temporal imputation problem, mitigating the error accumulation common in autoregressive methods, while SpecSTG~\citep{lin2024specstg} shifts the generation process to the spectral domain to better model systematic spatial patterns and improve computational efficiency. These models, while powerful, are typically specialized for ST graph-structured data and trained for a specific forecasting or imputation task.

A parallel and emerging trend is the development of LLM foundation models for time-series applications trained on vast, heterogeneous datasets for zero-shot forecasting. These models aim for generalization across diverse time-series without task-specific fine-tuning. Notable examples are Moirai-MoE~\citep{moiraimoe}, TimerXL~\citep{liu2025timerxllongcontexttransformersunified},  and Chronos~\citep{ansari2024chronoslearninglanguagetime}, which challenge the prevailing reliance on human-defined heuristics like frequency-based specialization. Instead of using separate modules for different time-series frequencies, Moirai-MoE~\citep{moiraimoe} incorporates a sparse Mixture of Experts (MoE)~\citep{Tresp2001} layer within its Transformer architecture. This design enables automatic, token-level specialization, allowing the model to dynamically capture a wide array of patterns and non-stationarities inherent in diverse time-series data. This represents a shift from designing specialized architectures for specific data structures (like graphs) to building more general, adaptable models that learn to handle heterogeneity internally.


\newpage
\section*{NeurIPS Paper Checklist}

\begin{enumerate}

\item {\bf Claims}
    \item[] Question: Do the main claims made in the abstract and introduction accurately reflect the paper's contributions and scope?
    \item[] Answer: \answerYes{} 
    \item[] Justification: We clarify the contributions and the scope in abstract and the introduction.

\item {\bf Limitations}
    \item[] Question: Does the paper discuss the limitations of the work performed by the authors?
    \item[] Answer: \answerYes{} 
    \item[] Justification: It has been discussed in Section~\ref{conc}.

\item {\bf Theory assumptions and proofs}
    \item[] Question: For each theoretical result, does the paper provide the full set of assumptions and a complete (and correct) proof?
    \item[] Answer: \answerNA{} 
    \item[] Justification: the paper does not include theoretical results

    \item {\bf Experimental result reproducibility}
    \item[] Question: Does the paper fully disclose all the information needed to reproduce the main experimental results of the paper to the extent that it affects the main claims and/or conclusions of the paper (regardless of whether the code and data are provided or not)?
    \item[] Answer: \answerYes{} 
    \item[] Justification: We provide details in Section~\ref{results}.

\item {\bf Open access to data and code}
    \item[] Question: Does the paper provide open access to the data and code, with sufficient instructions to faithfully reproduce the main experimental results, as described in supplemental material?
    \item[] Answer: \answerYes{} 
    \item[] Justification: We provide the data and code in supplementary materials.

\item {\bf Experimental setting/details}
    \item[] Question: Does the paper specify all the training and test details (e.g., data splits, hyperparameters, how they were chosen, type of optimizer, etc.) necessary to understand the results?
    \item[] Answer: \answerYes{} 
    \item[] Justification: We provide details in Section~\ref{results}.

\item {\bf Experiment statistical significance}
    \item[] Question: Does the paper report error bars suitably and correctly defined or other appropriate information about the statistical significance of the experiments?
    \item[] Answer: \answerYes{} 
    \item[] Justification: We provide details in Section~\ref{results}.

\item {\bf Experiments compute resources}
    \item[] Question: For each experiment, does the paper provide sufficient information on the computer resources (type of compute workers, memory, time of execution) needed to reproduce the experiments?
    \item[] Answer: \answerYes{} 
    \item[] Justification: We provide details in Section~\ref{results}.
    
\item {\bf Code of ethics}
    \item[] Question: Does the research conducted in the paper conform, in every respect, with the NeurIPS Code of Ethics \url{https://neurips.cc/public/EthicsGuidelines}?
    \item[] Answer: \answerYes{} 
    \item[] Justification: This paper conforms, in every respect, with the NeurIPS Code of Ethics.

\item {\bf Broader impacts}
    \item[] Question: Does the paper discuss both potential positive societal impacts and negative societal impacts of the work performed?
    \item[] Answer: \answerNA{} 
    \item[] Justification: We think our work will not have a significant social impact.
    
\item {\bf Safeguards}
    \item[] Question: Does the paper describe safeguards that have been put in place for responsible release of data or models that have a high risk for misuse (e.g., pretrained language models, image generators, or scraped datasets)?
    \item[] Answer: \answerNA{} 
    \item[] Justification: The paper poses no such risks.

\item {\bf Licenses for existing assets}
    \item[] Question: Are the creators or original owners of assets (e.g., code, data, models), used in the paper, properly credited and are the license and terms of use explicitly mentioned and properly respected?
    \item[] Answer: \answerYes{} 
    \item[] Justification: All assets used in this paper are properly credited.

\item {\bf New assets}
    \item[] Question: Are new assets introduced in the paper well documented and is the documentation provided alongside the assets?
    \item[] Answer: \answerYes{} 
    \item[] Justification: We provide the documentation in the code repo.

\item {\bf Crowdsourcing and research with human subjects}
    \item[] Question: For crowdsourcing experiments and research with human subjects, does the paper include the full text of instructions given to participants and screenshots, if applicable, as well as details about compensation (if any)? 
    \item[] Answer: \answerNA{} 
    \item[] Justification: The paper does not involve crowdsourcing nor research with human subjects.

\item {\bf Institutional review board (IRB) approvals or equivalent for research with human subjects}
    \item[] Question: Does the paper describe potential risks incurred by study participants, whether such risks were disclosed to the subjects, and whether Institutional Review Board (IRB) approvals (or an equivalent approval/review based on the requirements of your country or institution) were obtained?
    \item[] Answer: \answerNA{} 
    \item[] Justification: The paper does not involve crowdsourcing nor research with human subjects.

\item {\bf Declaration of LLM usage}
    \item[] Question: Does the paper describe the usage of LLMs if it is an important, original, or non-standard component of the core methods in this research? Note that if the LLM is used only for writing, editing, or formatting purposes and does not impact the core methodology, scientific rigorousness, or originality of the research, declaration is not required.
    \item[] Answer: \answerNA{} 
    \item[] Justification: The core method development in this research does not involve LLMs as any
important, original, or non-standard components.

\end{enumerate}

\end{document}